\documentclass[letterpaper, 10 pt, conference]{ICRA/ieeeconf}  % Comment this line out if you need a4paper

\IEEEoverridecommandlockouts                              % This command is only needed if 
\usepackage{graphics} % for pdf, bitmapped graphics files
\usepackage{graphicx}

\usepackage{amsmath} % assumes amsmath package installed
\usepackage{amsfonts}
\usepackage{subcaption}
\usepackage{float} 
\usepackage{booktabs}
\usepackage{multirow}
\usepackage{bm}
\usepackage[linesnumbered,ruled,vlined]{algorithm2e}
\usepackage[hidelinks]{hyperref}
\usepackage{amssymb}  % assumes amsmath package installed
\usepackage{cite}
\title{\LARGE \bf
Adaptive Diffusion Constrained Sampling for\\ Bimanual Robot Manipulation
}

\author{
  Haolei Tong\textsuperscript{1, *}, Yuezhe Zhang\textsuperscript{1, *}, Sophie Lueth\textsuperscript{1}, Georgia Chalvatzaki\textsuperscript{1, 2, 3}
  \thanks{*This project has received funding from the European Union’s Horizon Europe programme under Grant Agreement No. 101120823, project MANiBOT. \textsuperscript{1}TU Darmstadt, \textsuperscript{2}Hessian.AI, \textsuperscript{3}Robotics Institute Germany. \textsuperscript{*} indicates equal contribution. Email: \href{mailto:haoleitong24@gmail.com}{haoleitong24@gmail.com}
}
}

\begin{document}

\maketitle
\thispagestyle{empty}
\pagestyle{empty}

\begin{abstract}

Coordinated multi-arm manipulation requires satisfying multiple simultaneous geometric constraints across high-dimensional configuration spaces, which poses a significant challenge for traditional planning and control methods. In this work, we propose Adaptive Diffusion Constrained Sampling (ADCS), a generative framework that flexibly integrates both equality (e.g., relative and absolute pose constraints) and structured inequality constraints (e.g., proximity to object surfaces) into an energy-based diffusion model. Equality constraints are modeled using dedicated energy networks trained on pose differences in the Lie algebra space, while inequality constraints are represented via Signed Distance Functions (SDFs) and encoded into learned constraint embeddings, allowing the model to reason about complex spatial regions. A key innovation of our method is a Transformer-based architecture that learns to weigh constraint-specific energy functions at inference time, enabling flexible and context-aware constraint integration. Moreover, we adopt a two-stage batch-wise sampling strategy that improves precision and sample diversity by combining Langevin dynamics with resampling and density-aware re-weighting. Experimental results on dual-arm manipulation tasks show that ADCS significantly improves sample diversity and generalization in settings demanding precise coordination and adaptive constraint handling.  Our website is made publicly available at: \href{https://thomasston.github.io/ADCS.github.io/}{thomasston.github.io/ADCS.github.io/}

\end{abstract}
\section{Introduction}
Robots operating in human environments must be capable of performing increasingly complex tasks that involve interaction with diverse objects, unstructured scenes, and coordination across multiple end-effectors. As tasks grow in complexity~\cite{canny1988complexity}, such as moving large furniture, assembling components, individual manipulators often reach their physical and functional limits. Multidegrees-of-freedom systems, such as bimanual robots and mobile manipulators, provide a scalable way to address such limitations by enabling spatial coordination and collaborative manipulation~\cite{khatib1999mobile}.

One core capability that these systems unlock is collaborative transport, where multiple agents work together to move objects that are too large, heavy, or geometrically constrained for a single robot to handle. However, this capability comes at the cost of dramatically increased complexity: planning motion trajectories for such systems requires satisfying a wide range of spatial and physical constraints,e.g., maintaining stable grasps, ensuring collision-free motion among robots and with the environment, and respecting the kinematic limits of each agent. This results in high-dimensional problems with complex nonlinear constraints in configuration and task spaces~\cite{kingston2018sampling}.

Traditional constrained optimization methods and sampling techniques such as Markov Chain Monte Carlo (MCMC)~\cite{toussaint2024nlp} offer principled approaches to constraint satisfaction in such settings. However, their reliance on local gradient information often limits their capacity to explore multi-modal distributions or reason globally about the scene geometry. Recent generative approaches have shown promise; diffusion models were applied to $\mathrm{SE}(3)$ grasp synthesis~\cite{urain2023se} and to compositional object placement~\cite{yang2023compositional}, but they typically assume fixed constraint structures and manually specified weighting schemes. Moreover, they often struggle to scale to systems with high-dimensional, interdependent constraints, such as those in multi-arm or mobile manipulation platforms.

To address these limitations, we introduce \textbf{Adaptive Diffusion Constrained Sampling (ADCS)}, a generative framework for high-dimensional, constraint-aware sampling in multi-DoF robotic systems. ADCS is designed to integrate both equality and inequality constraints in $\mathrm{SE}(3)$ task space while remaining adaptable to different robot morphologies and task conditions. The key novelty lies in two adaptive mechanisms.
First, we introduce an \emph{Adaptive Constrained Conditioning}, in which the model during training learns to generate feasible $\mathrm{SE}(3)$ poses conditioned on the scene geometry via Signed Distance Fields (SDFs). During sampling, we apply differentiable constraints in joint space using a chain rule, enabling flexible post-hoc adaptation to robot-specific constraints such as joint limits, reachability, and grasp feasibility. Second, we propose a \emph{Compositional Weighting Transformer (CWT)} method, in which, instead of manually specifying energy-function weights, we incorporate a Transformer-based architecture that learns to dynamically compose task-specific energy terms, allowing the system to prioritize constraints adaptively across varying tasks and contexts, to allow for optimizing challenging tasks with varied co-occurring constraints. Moreover, the CWT can handle new types of constraints without requiring retraining.

In addition, ADCS uses a two-stage sampling strategy that combines Langevin dynamics, density-aware re-weighting, and Gauss-Newton refinement, which leads to faster convergence and better precision in constrained sampling. We evaluate ADCS in several collaborative manipulation scenarios, including object transport and pattern stippling, using both simulated and real-world multi-arm systems -- a two Franka Emika Panda arm system and a bimanual TIAGo robot. Across these settings, ADCS consistently outperforms baseline approaches in terms of task success, sampling efficiency, generalization to novel scenes, and constraint satisfaction.

To sum up, our main contributions are as follows. \textbf{(i)} We propose \emph{ADCS}, a diffusion-based framework for constraint-aware generative sampling in multi-DoF robotic systems.
\textbf{(ii)} We introduce CWT, a Transformer-based mechanism that dynamically composes constraint energies, eliminating manual weight tuning and generalizing to new constraints.
\textbf{(iii)} We formulate a two-stage batch-wise sampling process that enables the application of task- and robot-specific constraints in joint space, even post training.
\section{Related Works}
\textbf{Constrained Sampling.} \hspace{0.3em} Sampling under constraints is central to problems in MCMC, optimization, and robotics. With equality constraints, the task reduces to sampling on a manifold~\cite{suh2011tangent, kim2016tangent}, while linear inequalities lead to polytope sampling~\cite{chen2018fast}. \cite{toussaint2024nlp} proposed a restart-based NLP sampler for nonlinear constraints in robotics, assuming only local access to cost functions. In contrast, we focus on sampling from complex, multimodal constrained distributions.

\textbf{Generative Models for Constrained Optimization.} \hspace{0.3em} Deep generative models such as DDPM~\cite{ho2020denoising} and SVGD~\cite{liu2016stein} are powerful tools for sampling from multimodal distributions. Their ability to capture disconnected regions makes them attractive for constraint-aware inference~\cite{urain2024deep}. Existing works~\cite{ortiz2022structured}\cite{ yang2023compositional} apply diffusion models to constraint-aware tasks, but often focus on low-dimensional object poses or categorical approximations. Diffusion-based motion planners~\cite{carvalho2023motion, kurtz2024equality, pan2024model} reframe planning as sampling in energy-based models, but struggle to scale with multiple, nonlinear constraints. Our method addresses this by learning a compositional energy model with transformer-based adaptive weighting for rich constraint integration.

\textbf{Automatic Weight Tuning.} \hspace{0.3em} Designing appropriate constraint weights is tedious and sensitive~\cite{urain2023se}\cite{zhang2024multi}. Prior methods either optimize weights~\cite{kumar2021diffloop}\cite{cheng2024difftune} or learn them from data~\cite{marco2016automatic}\cite{loquercio2022autotune}, but these are not directly applicable to diffusion-based models. Recent approaches~\cite{urain2023se}\cite{yang2023compositional},\cite{carvalho2023motion} use fixed weights, whereas our method trains a transformer (CWT) to reason over constraint types and adaptively weight them at inference.

Compared to the prior work, Diffusion-CCSP~\cite{yang2023compositional}, it relies on fixed weights to compose constraints and utilizes basic Langevin dynamics for sampling. In contrast, our method incorporates the task-specific feature functions from ADCS, which are essential for enhancing generalization. And also our method benefits from transformer-based learnable constraint weighting and a more expressive two-stage sampling strategy, resulting in better performance across tasks.

% our method (ADCS) introduces a Transformer-based, dynamic, and learnable constraint weighting mechanism, along with a more expressive two-stage sampling strategy. This allows ADCS to better adapt to diverse constrained problem structures and sample effectively from complex, high-dimensional constraint manifolds, especially for structured, high-dimensional robotic tasks, thereby extending the scope of compositional constraint satisfaction.
\section{Preliminaries}
Here, we formally define the constrained sampling problem, describe the types of constraints considered, and outline the NLP-based sampling method used to generate training data for our model.

\textbf{Problem Statement.} \hspace{0.3em}~~~We consider a multi-DoF robotic system composed of multiple manipulators operating in a shared workspace. Let \( \mathcal{Q} \subset \mathbb{R}^{n\times d} \) denote the joint configuration space, where $n$ denotes the number of robots/end-effectors and $d$ denotes the Dof of each robot, and let \( \mathcal{C} = \{ c_i \} \) be a set of task-specific constraints on configurations \( q \in \mathcal{Q} \). Each constraint is either an equality \( c_i(q) = 0 \) or an inequality \( c_i(q) \leq 0 \). These constraints capture spatial relationships, kinematic feasibility, and task semantics across the robot and the environment.

\textbf{Constraint Types.} \hspace{0.3em} 
We consider the following types of constraints: \textbf{(i) $\mathrm{\textbf{SE(3)}}$ Pose Constraints:} enforce absolute or relative poses between end-effectors. Given two poses \( H_1, H_2 \in \mathrm{SE}(3) \), the constraint is \( H_1^{-1} H_2 = T_{\text{target}} \); \textbf{(ii) Orientation Constraints:} constrain either a specific axis (e.g., Y-axis of the gripper aligned with Z-axis of the world) or the full rotation \( R = R_{\text{goal}} \); \textbf{(iii) Midpoint Constraints:} enforce the midpoint between two end-effectors to match a target (equality) or lie on a surface (inequality); \textbf{(iv) Signed Distance Constraints:} encode proximity to object surfaces for surface contact or collision avoidance; \textbf{(v) Joint and Self-Collision Constraints:} enforce joint limits and prevent inter-link collisions using a set of differentiable bounding spheres.

\textbf{NLP Sampling.} \hspace{0.3em}
To generate training data, we use \textit{NLP Sampling}~\cite{toussaint2024nlp}, a method for sampling from the constraint-satisfying region defined by \( \mathcal{C} \). Each constraint \( c_i(q) \) is mapped to a slack term: inequality constraints use \( [c_i(q)]_+ \), and equality constraints use \( |c_i(q)| \), where \( [\cdot]_+ \) denotes the ReLU function. The aggregated slack vector \( s(q) \) defines the relaxed energy:
\begin{equation}
F_{\gamma, \mu}(q) = \gamma f(q) + \mu \| s(q) \|^2,
\end{equation}
where \( f(q) \) is a task-specific density and \( \gamma, \mu \) balance task likelihood and constraint satisfaction. Sampling proceeds in two stages: first, Gauss-Newton descent minimizes \( \|s(q)\|^2 \) over \( K_{\text{down}} \) steps to find a feasible configuration; then, interior sampling, such as manifold-RRT~\cite{suh2011tangent} or Langevin dynamics~\cite{ho2020denoising}, is applied for \( K_{\text{burn}} \) iterations to generate diverse samples from the feasible region. This approach produces high-quality training data for constraint-aware generative modeling.

\section{Adaptive Diffusion Constrained Sampling}
In this section, we describe our framework for generating diverse, constraint-satisfying solutions in multi-DoF robotic systems. We first introduce our compositional energy-based formulation for modeling constraints in a diffusion framework. We then describe the model architecture, training loss, and how the learned model can be used for efficient sampling at inference time.

\subsection{Adaptive Compositional Diffusion with Conditioning}
Given a set of constraints \( \mathcal{C} = \{c_i\} \), we aim to sample robot poses \( H \in \mathrm{SE}(3)^n \) (for \( n \) end-effectors) that satisfy all constraints in \( \mathcal{C} \). We model the conditional distribution over poses using an energy-based model (EBM), enabled by the use of diffusion models. Unlike standard DDPMs~\cite{ho2020denoising} that generate samples from noise via learned denoisers, we use diffusion training solely to learn a score function for constraint-driven energy modeling based on denoising score matching~\cite{song2020improved}. For each individual constraint \( c \in \mathcal{C} \), we define an energy function \( E_\theta(H \mid c) \), and the associated probability is given by \(p(H \mid c) \propto \exp\bigl(-E_\theta(H \mid c)\bigr) \). We assume these models assign approximately uniform mass over the feasible region of each constraint. To satisfy all constraints jointly, we construct a composed distribution by minimizing the sum of individual energy functions
\begin{equation}
H_0 = \arg\min_H \sum_{c \in \mathcal{C}} E_\theta(H \mid c).
\end{equation}
Similarly to~\cite{liu2022compositional}, the joint diffusion distribution over noisy latent variables \( H_k \) (at timestep \( k \)) under all constraints \( c_0, \ldots, c_N \in \mathcal{C} \) is given by
\begin{equation}
p(H_k \mid c_0, \ldots, c_n) \propto p(H_k) \prod_{i=0}^N \frac{p(H_k \mid c_i)}{p(H_k)},
\end{equation}
which leads to the following form for the composed energy
\begin{equation}
\begin{aligned}
\label{equ:compose}
E_\theta(H_k \mid c_0, \ldots, c_n) = w_0 E_\theta(H_k \mid c_0)\\
+ \sum_{i=1}^N w_i \left( E_\theta(H_k \mid c_0, c_i) - E_\theta(H_k \mid c_0) \right),
\end{aligned}
\end{equation}
where \( c_0 \) is a fundamental constraint (e.g., task-space goal, which should be satisfied in every task), and the remaining energies are conditioned on the satisfaction of \( c_0 \), since in all of our tasks, the EEF’s relative pose constraints \(c_0\) should be satisfied. The weights \( w_i \) determine the influence of each constraint during composition and can be either fixed or learned.

To enable learning in the Lie group \( \mathrm{SE}(3)^n \), we inject Gaussian noise in the Lie algebra \( \mathfrak{se}(3)^n \), to maintain manifold consistency. Specifically, for a clean pose \( H \), we apply a perturbation via the exponential map~\cite{urain2023se}:
\begin{equation}
\widehat{H} = H \cdot \text{Expmap}(\epsilon), \quad \epsilon \sim \mathcal{N}(0, \sigma_k^2 I), \quad \epsilon \in \mathbb{R}^{6n},
\end{equation}
where \( \sigma_k \) is the noise scale at timestep \( k \), and Expmap maps noise vectors in the tangent space to elements of the Lie group. To train the model, we adopt a denoising score matching objective~\cite{song2020improved}, which minimizes the discrepancy between the model's gradient and the gradient of the Gaussian perturbation distribution. The training loss is
\begin{equation}\label{equ:lossfunc}
\scalebox{0.78}{$
\mathcal{L} = \frac{1}{L} \sum_{k=0}^{L}
\mathbb{E}_{H, \widehat{H}} \left[
\left\|
\nabla_{\widehat{H}} E_\theta(\widehat{H} \mid \mathcal{C}, k)
-
\nabla_{\widehat{H}} \log q(\widehat{H} \mid H, \sigma_k^2 I)
\right\|^2
\right]
$}
\end{equation}
where \( q(\widehat{H} \mid H, \sigma_k^2 I) \) is the Gaussian perturbation distribution in the Lie algebra \( \mathfrak{se}(3)^n \), and \( \nabla_{\widehat{H}} \) denotes the gradient w.r.t. the perturbed pose \( \widehat{H} \in \mathrm{SE}(3)^n \). Since \( \mathrm{SE}(3)^n \) is a Lie group, gradients are computed in the associated tangent space via a local reparameterization (details in the Appendix).
\\
This formulation learns an EBM implicitly through a diffusion process. While standard diffusion models learn a generative process via denoising predictions, our framework instead learns the score function, i.e., the gradient of the log-probability, directly. The trained energy function \( E_\theta(\widehat{H} \mid \mathcal{C}, k) \) thus estimates the gradient of the log-density at different noise levels.
At inference time, we discard the forward diffusion and use the learned score function as input to Langevin dynamics, iteratively refining a sample by ascending the composed energy landscape.

In addition, we introduce an \emph{Adaptive Constrained Conditioning} mechanism to bridge robot-level and task-level constraints. During training, task-level conditions are incorporated into the energy network via a FiLM-based modulation mechanism. For SDF constraints, we jointly train a network that predicts signed distance values from point cloud inputs. At inference time, gradients are propagated through forward kinematics using the chain rule, which enables flexible enforcement of robot-specific constraints (e.g., joint limits and collision) without retraining and ensures adaptability across both robot and task-level constraints.

% as we describe in~Sec.~[TODO]

\begin{figure*}[t]
  \centering
  \vspace{0.2cm}
  \includegraphics[width=\textwidth]{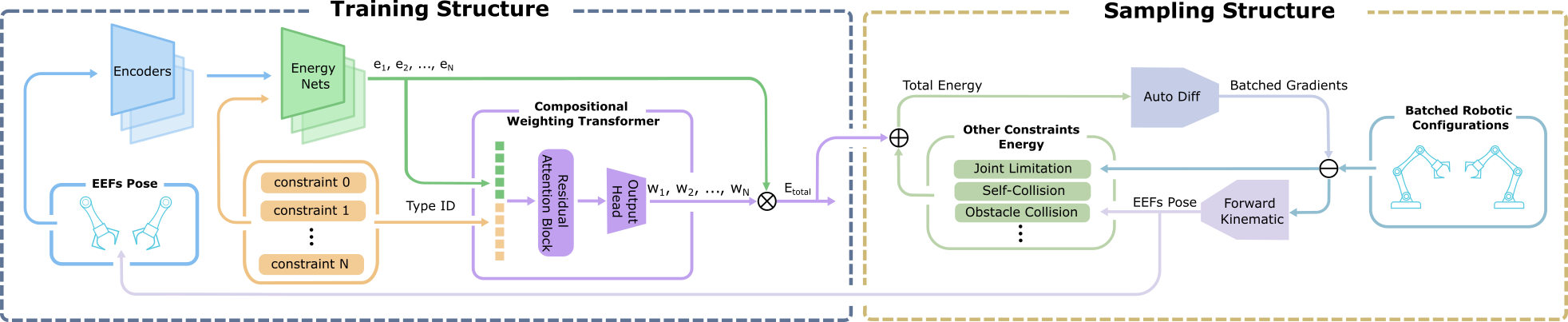}
  \vspace{-0.4cm}
  \caption{Overview of the Adaptive Diffusion Constrained Sampling (ADCS) architecture. The model consists of three components: a constraint feature encoder, an energy network (MLP), and the Compositional Weighting Transformer (CWT). During training, noisy poses in \(\mathrm{SE}(3)^n\) are used to learn constraint-aware energy functions. In inference, sampling is performed in batched joint space using Langevin dynamics guided by the learned energy landscape.}
  \label{fig:train net}
  \vspace{-0.6cm}

\end{figure*}

\subsection{Compositional Weighting Transformer}
To enable flexible and context-aware integration of multiple constraint energies, we introduce CWT. Rather than statically assigning scalar weights to each constraint model, the CWT learns dynamic composition weights conditioned on the current energy values and their interrelations. 
The key intuition is that the Transformer~\cite{vaswani2017attention} enables\emph{ contextual weighting}: it attends to all constraint-specific energy values jointly and can adapt the relative importance of each constraint depending on their magnitudes, types, or combinations. This makes the composition mechanism \emph{permutation-invariant} (no fixed order of constraints is required) and \emph{adaptive to task conditions}, as the model can, for instance, downweight a nearly satisfied constraint and prioritize violated ones during inference.

As illustrated in Figure~\ref{fig:train net}, each energy scalar \( E_{\theta}(\widehat{H} \mid c_i) \) for constraint \( c_i \in \mathcal{C} \), together with the constraints identifier, is treated as a token. Since each type of constraint has its own independent energy encoder, all constraint types are also implicitly encoded. To handle unseen constraints during inference, we also introduce a special UNK identifier, allowing the model to generalize beyond the training set. These tokens are concatenated and passed through a Transformer, which outputs a set of normalized weights \( \{w_i\} \). The final composed energy is $E_{\text{total}}(\widehat{H}) = \sum_{i} w_i E_{\theta}(\widehat{H} \mid c_i)$. We also use positional encodings for the input tokens, but unlike the original Transformer~\cite{vaswani2017attention}, in CWT, those are randomly sampled at each iteration since there is no intrinsic ordering among constraints. This design supports modularity and compositional generalization across tasks and scenes without retraining. Compared with approaches that fuse constraint features directly via fully connected layers~\cite{ha2016hypernetworks,jia2016dynamic}, CWT preserves the identity of each constraint’s energy and defers fusion to the composition stage.  This enables each constraint energy to preserve both geometric structure and semantics, while allowing adaptive trade-offs between them as the constraint energies change with the environment.
CWT is trained jointly with the energy networks and feature encoders using the diffusion-based loss in Eq.~\ref{equ:lossfunc}, without requiring additional supervision. 
% A full training pipeline and implementation details are provided in \ref{sec:append-train} and Algorithm~\ref{algo:train} in the Appendix.

\subsection{Two-stage Constrained Sampling} 
\label{sec:sampling}

At inference time, we use the learned diffusion model to generate batched joint configurations that satisfy a set of geometric constraints, where the batch size is the number of drawn samples. To this end, we adopt a two-stage sampling strategy inspired by~\cite{toussaint2024nlp}, which combines Annealed Langevin Monte Carlo (ALMC) sampling with resampling and correction steps, as shown in Algo.~\ref{alg:seqsampling}.

\textbf{Annealed Langevin Sampling.} \hspace{0.3em} In each ALMC call, we first draw a batch of joint configurations \( q_L^{n_s} \) from a standard Gaussian distribution. Using PyTorch Kinematics~\cite{Zhong_PyTorch_Kinematics_2024}, we compute the corresponding end-effector poses \( H_L^{n_s} \in \mathrm{SE}(3)^n \) via forward kinematics. These poses are then passed through the energy network to compute their corresponding energies \( E_{\theta}(H_L^{n_s} \mid \mathcal{C}, k) \). Next, we compute the batched gradient of the energy w.r.t. the joint configuration \( q_k^{n_s} \) using PyTorch’s AutoDiff and vmap~\cite{paszke2019pytorch}. This gradient is obtained via the chain rule $
\nabla_{q} E_\theta = \frac{\partial E_\theta(H_k \mid \mathcal{C}, k)}{\partial H_k} \cdot \frac{\partial H_k}{\partial q_k^{n_s}}$. The joint configurations are then updated using a Langevin step. To improve convergence, we adopt a second-order strategy similar to~\cite{carvalho2023motion}, but instead of relying solely on first-order gradients, we introduce an approximate batch-wise Gauss-Newton update during this correction phase. The update direction is computed as
$g_{n_s} = -(J^\top J)^{-1} J^\top r$,
where \( J=\frac{\partial r}{\partial q}=\frac{\partial r}{\partial H}\frac{\partial H}{\partial q} \) and \(r\) is the residual value. This approximation accelerates convergence while preserving the structure of the learned data distribution. Importantly, we restrict this correction to the refinement phase to avoid distorting the learned distribution during stochastic sampling, as discussed in~\cite{girolami2011riemann,leimkuhler2013rational}.

% Following each ALMC sampling pass, we perform a deterministic \emph{second-order refinement} step to improve the quality and constraint satisfaction of the sampled configurations. In this phase, we reduce the step size and omit the Gaussian noise, allowing the model to converge more precisely. We apply a Gauss-Newton correction \( g_{n_s} \) during this step, using approximate second-order information to guide updates in joint space. Importantly, we restrict this correction to the refinement phase to avoid distorting the learned distribution during stochastic sampling, as discussed in~\cite{girolami2011riemann,leimkuhler2013rational}.

\textbf{Resampling Phase.} \hspace{0.3em} 
After the initial ALMC sampling and second-order refinement, we perform a resampling step to promote diversity among the generated configurations. We begin by sorting all samples based on their energy values and selecting a fixed number of the top-performing configurations. To avoid over-representing densely clustered solutions, we estimate the sample density \( \rho(x) \) using Kernel Density Estimation (KDE) with a Gaussian kernel~\cite{silverman2018density}:
\begin{equation}
\scalebox{0.9}{$
  \rho(x)
  = \frac{1}{M\,h}
    \sum_{i=1}^M
    K\left(\frac{x - x_i}{h}\right), 
  \quad
  K(u) = \frac{1}{\sqrt{2\pi}} \exp\left(-\frac{u^2}{2}\right),
  $}
\end{equation}
where \( M \) is the number of retained samples, \( x_i \) are the sampled configurations, and \( h \) is the kernel bandwidth. Replication weights are assigned inversely proportional to \( \rho(x_i) \): samples in dense regions are duplicated less frequently, while those in sparse regions are upweighted. This reweighting encourages broader exploration of the configuration space, improving the final sample diversity.

\begin{algorithm}[]
\caption{Adaptive Diffusion Constrained Sampler}\label{alg:seqsampling}
\KwIn{Hyper-Parameter set $H$, Constraint set $C$
}
\KwOut{Final batched samples $q_0^{n_s}$}
Initialize $q_L^{n_s} \sim p_L(q)$\;
$q_0^{n_s}, e_0^{n_s}\gets \text{ALMC}(q_L^{n_s},H,C)$ \tcp*[r]{ $\triangleright$ Algo.~\ref{alg:ALM}}
$q_0^{n_r} \gets \text{sort}(q_0^{n_s}, e_0^{n_s}$)\\
$\rho(q) \gets \text{KDE}(q_0^{n_r})$ \tcp*[r]{Estimate density}
$w_i \propto 1 / \rho(q_i)$ \tcp*[r]{Get the weights}
$q_L^{n_s} \gets \text{repeat}(q_0^{n_r}, w_i) $ \tcp*[r]{Repeat data}
$q_0^{n_s}, e_0^{n_s}\gets \text{ALMC}(q_L^{n_s},H,C)$ \tcp*[r]{Final \(q_0^{n_s}\)}

\end{algorithm}

\vspace{-0.6cm}

\begin{algorithm}[]
\caption{Annealed Langevin Markov Chain Monte Carlo Sampler (ALMC)}\label{alg:ALM}
\KwIn{$H$, $C$, initial batched samples $q_L^{n_s}$
}
\KwOut{Final batched samples $q_0^{n_s}$}
\For{$k \leftarrow L$ \KwTo $1$}{
  Compute forward kinematics $H_k$ and energy $E(H_k,k,C)$\;
  Select step size $\alpha_k$ and sample Gaussian noise $\xi$\;
  Update samples by Langevin step: \\
  \Indp
  $q_{k-1} \leftarrow q_k - \tfrac{\alpha_k^2}{2}\nabla E(H_k,k,C) + \alpha_k \xi$\;
  \Indm
}
\textbf{Final refinement:} Apply Langevin updates with fixed step $\alpha_f$ combined with Gauss-Newton refinement for a fixed numbers of iterations\;
\Return{$q_0^{n_s}$}
\end{algorithm}

\section{Experimental Results}
We evaluate our method, \textbf{ADCS}, across a suite of constrained motion generation tasks and compare it to representative baselines:
\textbf{\emph{Gauss–Newton}} uses the predefined sampling cost, with updates computed as \(\Delta q = -(J^\top J)^{-1} J^\top r\).  
\textbf{\emph{Diffusion-CCSP}}~\cite{yang2023compositional} follows a similar energy-based formulation, with fixed (uniform) constraint weights and annealed Langevin dynamics as the sampler.  
\textbf{\emph{NLP-Sampling}} uses the method from~\cite{toussaint2024nlp}, note that its internal optimization loop is not externally controllable, so we specify only the number of final samples. All methods are evaluated over 10 random seeds (0–9), and we report average metrics across these runs. For fair comparison, Gauss–Newton and CCSP are run for 600 iterations, matching the sampling budget of ADCS.

\subsection{Simulation Experiment}
\label{sec:task_def}
\begin{figure}[]            
  \centering
  \vspace{0.15cm}
  \includegraphics[width=2.4in]{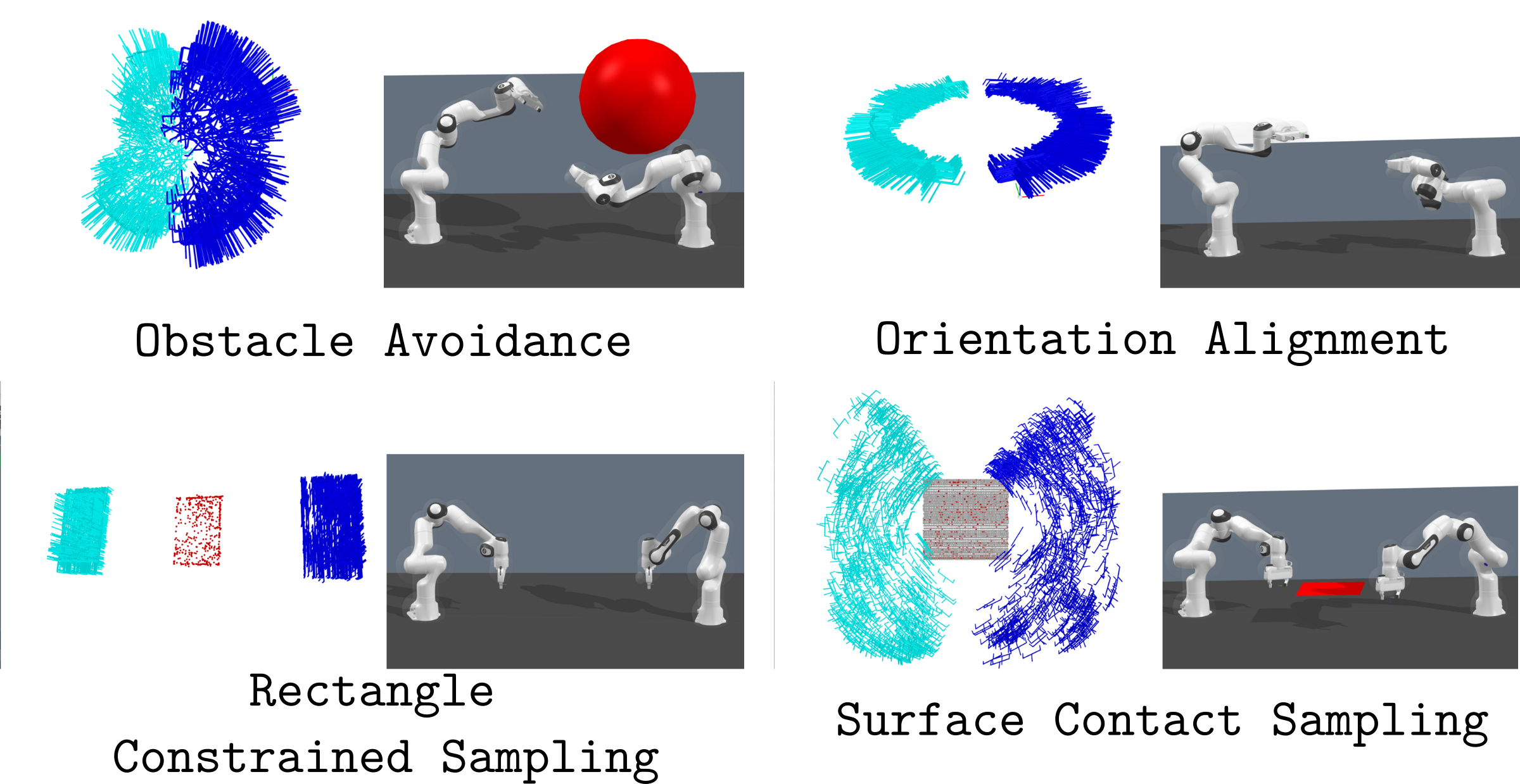}
  \vspace{-0.15cm}
  \caption{Sampled configurations for simulation tasks. Each image shows the robots' EE poses satisfying various constraints. Red dots indicate end-effector midpoints. Gray regions are constrained surfaces. Tasks increase in complexity from relative pose and midpoint constraints to full orientation and surface contact via SDFs.}
  \label{fig:sim_tasks}  
  \vspace{-0.6cm}
\end{figure}

We design four simulation tasks featuring increasingly complex geometric and pose constraints (see Figure \ref{fig:sim_tasks}). All tasks are performed using 2 Franka robots and visualized using the \href{https://marctoussaint.github.io/robotic}{Robotic} package. For each task, we generate 500 samples using each method (batch size is 500). 

All tasks involve a relative pose constraint between the two end-effectors and a midpoint constraint. \textbf{(1)} \emph{Obstacle Avoidance}: relative pose + symmetric midpoint constraint and adds a spherical obstacle to avoid.
\textbf{(2)} \emph{Orientation Alignment:} adds orientation alignment with the table plane without obstacle, i.e., the EEF’s Y-axis is aligned with the world’s Z-axis. \textbf{(3)} \emph{Rectangle Constrained Sampling:} midpoint lies inside a box (defined via bounds), with fixed EEF rotation matrix. \textbf{(4)} \emph{Surface Contact Sampling:} replaces the region with a surface defined by a point cloud and loosens the orientation constraint to be normal to the surface, i.e., the EEF’s Z-axis is aligned with the negative world Z-axis.

In Table~\ref{tab:all_compare}, we report \textbf{mean, median, and third-quartile values for position and rotation errors} across four tasks. Under a fixed sampling budget (600 iterations), ADCS consistently yields significantly lower constraint satisfaction errors than Gauss–Newton and Diffusion-CCSP. Especially in tasks with smaller feasible regions, such as Rectangle Constrained Sampling, our method demonstrates stronger performance compared to the baselines. And compared to the state-of-the-art NLP-Sampling, it outperforms in most metrics while requiring shorter sampling time. In contrast, ADCS offers strong accuracy with significantly improved efficiency and broader applicability.

We also compare the \textbf{spatial quality of samples} generated by NLP Sampling, CCSP, and ADCS in Table~\ref{tab: data_density_comparison}, evaluating both \textbf{coverage} and \textbf{uniformity} of the resulting distributions. Task \emph{Obstacle Avoidance} and \emph{Orientation Alignment} focus on the 3D position of a single end-effector, while Task \emph{Rectangle Constrained Sampling} and \emph{Surface Contact Sampling} assess the midpoint distribution between two arms. Coverage is computed as the fraction of occupied voxels in a predefined 3D grid, while uniformity is measured by the variance in density across voxels, lower variance indicates a more even spread of samples. As shown in the table, ADCS achieves superior coverage in most tasks and lower variance in most cases, highlighting its ability to explore constraint-satisfying regions more thoroughly.

% Due to the relatively coarse voxel grid used (see Appendix~\ref{sec:voxel_density} and \ref{sec:append_density_compare}), absolute coverage values are modest but still reflect comparative performance.

\begin{table}
\centering
\vspace{0.1cm}
\resizebox{0.45\textwidth}{!}{
\begin{tabular}{ccccc} 
\midrule
\multicolumn{5}{c}{Data Distribution Coverage}                                                                                                                                                                                                                                                        \\ 
\midrule
             & \begin{tabular}[c]{@{}c@{}}Obstacle \\Avoidance\end{tabular} & \begin{tabular}[c]{@{}c@{}}Orientation \\Alignment\end{tabular} & \begin{tabular}[c]{@{}c@{}}Rectangle \\Constrained Sampling \end{tabular}            & \begin{tabular}[c]{@{}c@{}}Surface\\ Contact Sampling\end{tabular}  \\ 
\midrule
NLP Sampling & \textbf{0.0321}                                              & 0.0235                                                          & 0.0425                                                                         & -                                                                    \\
CCSP         & 0.0310                                                       & 0.0342                                                          & 0.41                                                                           & 0.2407                                                               \\
ADCS (ours)  & 0.0315                                                       & \textbf{0.0362}                                                 & \textbf{0.6}                                                                   & \textbf{0.6425}                                                      \\ 
\midrule
\multicolumn{5}{c}{Data Distribution Uniformity(density variance) ~}                                                                                                                                                                                                                                                     \\ 
\midrule
NLP Sampling & 0.3388                                                       & 0.5260                                                          & 126.7000                                                                       & -                                                                    \\
CCSP         & \textbf{\textbf{0.1687}}                                     & 0.1548                                                          & 13.7969                                                                        & 5.0834                                                               \\
ADCS (ours)  & 0.3669                                                       & \textbf{\textbf{\textbf{\textbf{0.1338}}}}                      & \textbf{\textbf{\textbf{\textbf{\textbf{\textbf{\textbf{\textbf{1.9097}}}}}}}} & \textbf{\textbf{1.5696}}                                             \\
\bottomrule
\end{tabular}
}
\vspace{-0.15cm}
\caption{Comparison of \textbf{coverage as sampled} and \textbf{data uniformity} by NLP Sampling, CCSP and ADCS.}
\label{tab: data_density_comparison}
\vspace{-0.4cm}
\end{table}

\begin{figure}[]            
  \centering                   
  \includegraphics[width=2.5in]{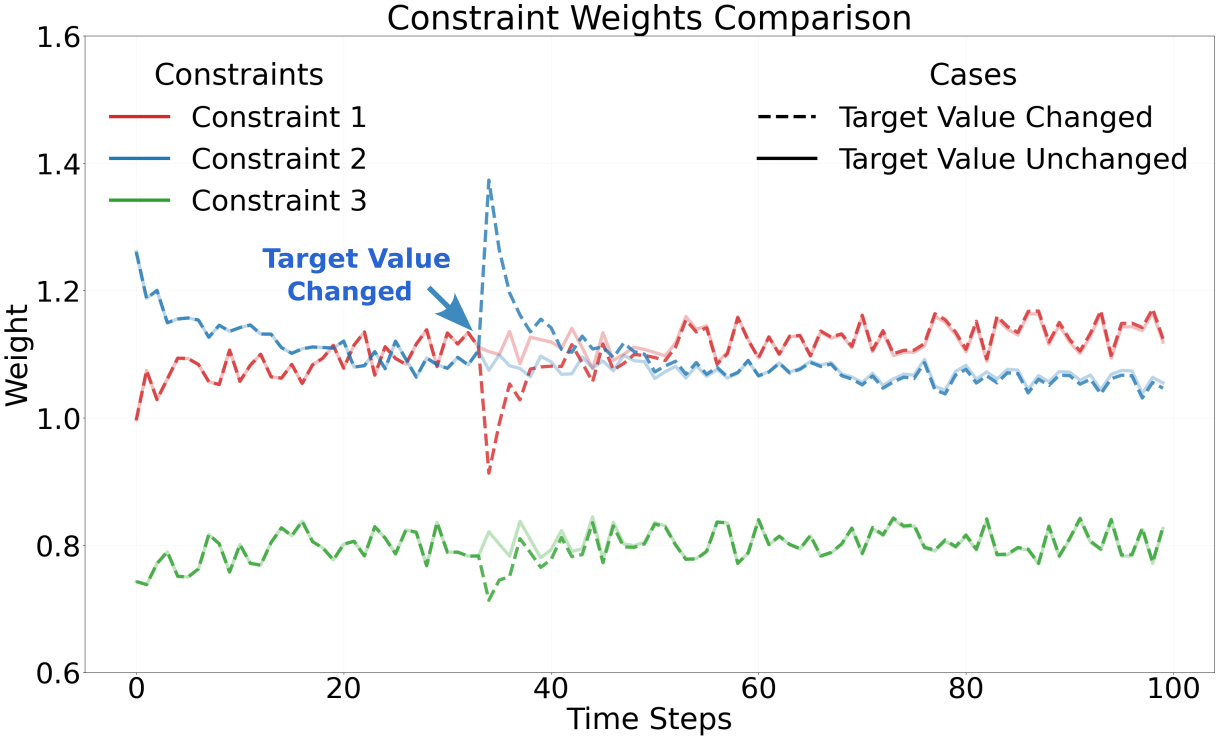}
  \vspace{-0.1cm}
  \caption{\textbf{Dynamic performance of CWT} with three different constraint weights under two scenarios: (i) changing target values during inference, and (ii) fixed target values.}
  \label{fig:dynamic_weights}  
  \vspace{-0.6cm}
\end{figure}

In order to validate the \textbf{dynamic performance of CWT}, we intentionally modified the target value of one constraint during inference, such as the desired relative pose between the two EEFs, which causes the sampled points to fall outside the feasible region of that constraint, thereby testing the model’s adaptability.  As illustrated in Figure~\ref{fig:dynamic_weights}, during inference, a change in the constraint target value leads to a corresponding adjustment of the weights, which enables the system to quickly recover back to the feasible region.

\begin{table*}[t]
\centering
\vspace{0.2cm}
    % \resizebox{\textwidth}{!}{
\begin{tabular}{c|cc|cc|cc|cc|c} 
\toprule
\multirow{2}{*}{} & \multicolumn{2}{c|}{RPE (mm)}          & \multicolumn{2}{c|}{MPE (mm)}              & \multicolumn{2}{c|}{RRE (rad)}                       & \multicolumn{2}{c|}{ERE (rad)}                       & Time (s)          \\
                  & Median          & Q3              & Median            & Q3                & Median                     & Q3                & Median                     & Q3                & Mean             \\ 
\midrule
\multicolumn{10}{c}{Obstacle Avoidance}                                                                                                                                                                                        \\ 
\midrule
NLP Sampling      & 0.0180          & 0.0660          & 0.0070            & 0.0292            & 4.652e-5                   & 0.0001            & -                          & -                 & 61.4200          \\
Gauss-Newton      & \textbf{0.0001} & 0.0003          & 7.649e-5          & 0.0001            & 1.718e-7                   & 3.215e-7          & -                          & -                 & \textbf{6.5329}  \\
CCSP              & 5.7985          & 23.6867         & 9.9820            & 31.7004           & 0.0128                     & 0.0373            & -                          & -                 & 9.1025           \\
ADCS (ours)       & \textbf{0.0001} & \textbf{0.0002} & \textbf{6.837e-5} & \textbf{8.652e-5} & \textbf{1.560e-7}          & \textbf{2.149e-7} & -                          & -                 & 10.3321          \\ 
\midrule
\multicolumn{10}{c}{Orientation Alignment}                                                                                                                                                                                        \\ 
\midrule
NLP Sampling      & 0.0358          & 0.0639          & 0.0177            & 0.0350            & 7.012e-5                   & 0.0001            & 0.0002                     & 0.0003            & 72.8657          \\
Gauss-Newton      & 13.5566         & 109.0678        & 2.1668            & 46.6404           & 0.0366                     & 0.3812            & 3.576e-07                  & 0.0185            & \textbf{7.6832}  \\
CCSP              & 12.5211         & 38.7049         & 11.1668           & 38.5522           & 0.0101                     & 0.0309            & 0.0341                     & 0.0671            & 9.3460           \\
ADCS (ours)       & \textbf{0.0050} & \textbf{0.0060} & \textbf{0.0026}   & \textbf{0.0029}   & \textbf{\textbf{1.495e-5}} & \textbf{1.611e-5} & \textbf{\textbf{1.192e-7}} & \textbf{1.788e-7} & 10.0325          \\ 
\midrule
\multicolumn{10}{c}{Rectangle Constrained Sampling}                                                                                                                                                                                        \\ 
\midrule
NLP Sampling      & 0.0176          & 0.0712          & 1.0019            & 4.4726            & 0.00003                    & 0.00009           & 0.00001                    & 0.00004           & 61.3578          \\
Gauss-Newton      & 0.0008          & 121.098         & 0.3174            & 9.4909            & 1.412e-06                  & 0.2704            & 5.277e-7                   & 3.418e-5          & \textbf{7.7406}  \\
CCSP              & 170.4515        & 257.2366        & 21.8248           & 56.5533           & 0.0458                     & 0.1007            & 0.1334                     & 0.2218            & 10.3574          \\
ADCS (ours)       & \textbf{0.0001} & \textbf{0.0002} & \textbf{0.1690}   & \textbf{0.2267}   & \textbf{\textbf{1.727e-7}} & \textbf{2.394e-7} & \textbf{1.334e-7}          & \textbf{1.883e-7} & 11.3247          \\ 
\midrule
\multicolumn{10}{c}{Surface Contact Sampling}                                                                                                                                                                                        \\ 
\midrule
NLP Sampling      & -               & -               & -                 & -                 & -                          & -                 & -                          & -                 & -                \\
Gauss-Newton      & 66.8540         & 247.1499        & 4.4389            & 44.8222           & 0.0875                     & 0.5902            & 0.0003                     & 0.0188            & \textbf{8.0048}  \\
CCSP              & 23.7275         & 58.6257         & 11.3875           & 23.6884           & 0.0282                     & 0.0691            & 0.0710                     & 0.1411            & 13.7951          \\
ADCS (ours)       & \textbf{0.0010} & \textbf{0.0013} & \textbf{1.8522}   & \textbf{2.8253}   & \textbf{7.439e-7}          & \textbf{9.614e-7} & \textbf{0}                 & \textbf{5.960e-8} & 14.9514          \\
\bottomrule
\end{tabular}
  % }
  \vspace{-0.1cm}
  \caption{\textbf{Comparison of Gauss–Newton, CCSP, and ADCS} across 4 tasks. For each task and method, we report \textbf{RPE} (relative position error (mm)), \textbf{MPE} (mid-point position error (mm)), \textbf{RRE} (relative rotation error (rad)), and \textbf{ERE} (end-effector rotation error in the world frame (rad)) by median and third quartile (Q3), and average computation time (s). }
    \label{tab:all_compare}
\vspace{-0.4cm}
\end{table*}

\subsection{Real-World Experiment}

\begin{figure}            
  \centering                   
  \includegraphics[width=2.5in]{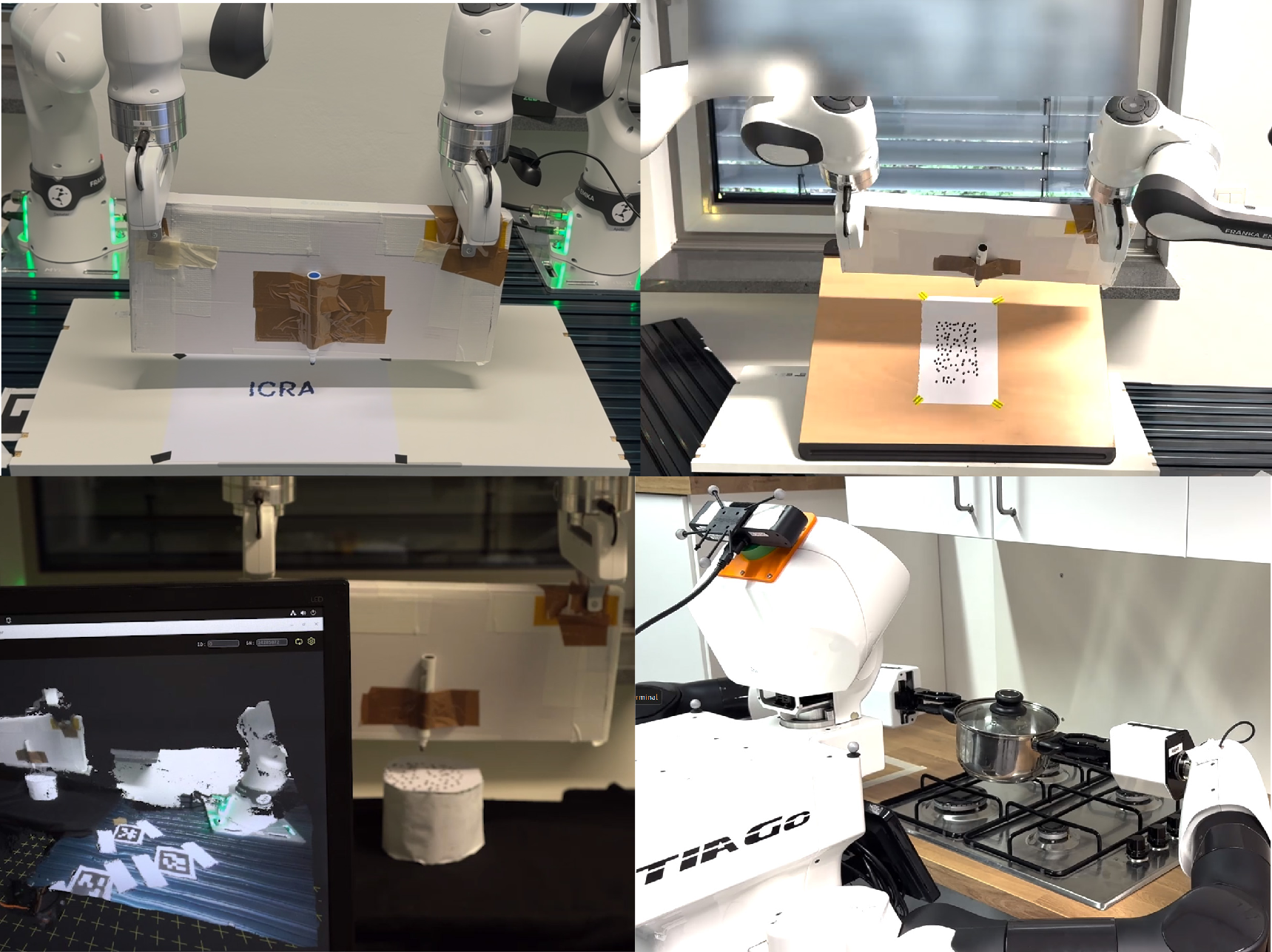}
  \vspace{-.1cm}
   \caption{The real-world experimental setup involves two Franka robots collaboratively grasping a pen for stippling operations and a TIAGo to carry objects using both hands. }
  \label{fig:expsetup}  
  \vspace{-.6cm}
\end{figure}

We have designed eight different tasks that integrate our ADCS with motion planning. Under various specified constraints, these eight tasks involve \href{https://vhartmann.com/robo-stippling/}{stippling} operations, i.e., two Franka robots collaboratively grasp a pen to perform different stippling tasks as shown in Figure~\ref{fig:expsetup}. In addition, we conducted tests on the TIAGo robot, allowing it to carry objects using both hands.

To integrate with motion planning, we first perform radius-based thinning on the original sampled points to filter out all points within a specified radius of the selected points. We then sort the remaining points using different, task-specific sampling and sorting strategies. After sorting, we inspect the distance between consecutive points in the joint space. Whenever the distance exceeds a preset threshold, we use the Gauss-Newton~\cite{Gauss1857} method to project to obtain a closer joint configuration, which usually converges in just two to three iterations.

% In order to make the sampled points smoother in the joint space, we enforce temporal smoothness by penalizing large jumps between consecutive configurations in the inference phase,

% \begin{equation}
% % \scalebox{0.9}{$
%     \Delta \bm{q}
% = 
% \begin{bmatrix}
% q_2 - q_1\\
% q_3 - q_2\\
% \vdots\\
% q_{n_s} - q_{n_s-1}
% \end{bmatrix} \quad 
% L_{\mathrm{cont}}
% = \sum_{n=2}^{n_s} \big\|\Delta \bm{q}_n\big\|_2^2,
% % $}
% \end{equation}
% where \(q_n\) is the joint configuration of sampled points, and the \(L_{cont}\) is the penalty function for smoothness. 
% After sorting, we inspect the distance between consecutive points in the joint space. Whenever the distance exceeds a preset threshold, we use the Gauss-Newton~\cite{Gauss1857} method to project to obtain a closer joint configuration, which usually converges in just two to three iterations. 

\begin{figure*}[t]            
  \centering
  \vspace{0.15cm}
  \includegraphics[width=.95\textwidth]{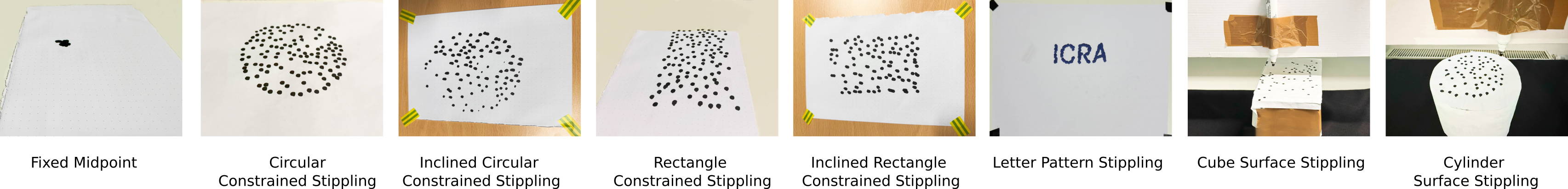}
  \vspace{-0.15cm}
  \caption{Final stippling layout of 8 different real-world tasks.}
  \label{fig:exp-layout}  
  \vspace{-0.5cm}
\end{figure*}

As illustrated in Figure~\ref{fig:exp-layout}, we evaluate eight real-world tasks. All tasks involve a relative pose constraint between the two end-effectors, as well as orientation constraints for the end-effectors. \textbf{(1)} \emph{Fixed Midpoint with Orientation Alignment}: constrains the midpoint between the two arms to a fixed location. \textbf{(2)} \emph{Circular Constrained Stippling} and \textbf{(3)} \emph{Inclined Circular Constrained Stippling}: restrict the midpoint to lie within a (inclined) circular bounding box. \textbf{(4)} \emph{Rectangle Constrained Stippling} and \textbf{(5)} \emph{Inclined Rectangle Constrained Stippling}: restrict the midpoint to lie within a (inclined) rectangle bounding box. \textbf{(6)} \emph{Letter Pattern Stippling} requires the midpoint to follow point clouds generated from mesh models of letters. Finally, \textbf{(7)} \emph{Cube Surface Stippling} and \textbf{(8)} \emph{Cylinder Surface Stippling}: constrain the midpoint to the top surface of real objects, with point clouds obtained directly from a camera.

In Table~\ref{tab: appendix-realexp-result}, we present the constraint error and the ratio of valid points for each task. Note that each task was tested ten times, and in every task, 500 points were generated through sampling. We consider a point valid if its positional error is less than 3 mm and its orientational error is less than 0.005 rad. As shown, the probability of valid points exceeds 80\% in most tasks.

\begin{table}
\centering
\resizebox{0.45\textwidth}{!}{
\begin{tabular}{cccccc} 
\toprule
                                                                                    & RPE    & MPE    & RRE    & ERE      & VPR    \\ 
\midrule
\begin{tabular}[c]{@{}c@{}}Fixed Midpoint with \\Orientation Alignment\end{tabular} & 6.9647 & 1.6872 & 0.0074 & 0.0042   & 72.60  \\ 
\midrule
\begin{tabular}[c]{@{}c@{}}Circular \\ConstrainedStippling\end{tabular}             & 0.0706 & 0.0024 & 0.0001 & 4.27e-05 & 98.40  \\ 
\midrule
\begin{tabular}[c]{@{}c@{}}Inclined Circular \\Constrained Stippling\end{tabular}   & 0.2818 & 0.0018 & 0.0003 & 0.0001   & 88.40  \\ 
\midrule
\begin{tabular}[c]{@{}c@{}}Rectangle \\Constrained Stippling\end{tabular}           & 0.1541 & 0.0097 & 0.0002 & 0.0002   & 90.60  \\ 
\midrule
\begin{tabular}[c]{@{}c@{}}Inclined Rectangle \\Constrained Stippling\end{tabular}  & 0.6734 & 0.0312 & 0.0006 & 0.0007   & 84.80  \\ 
\midrule
Letter Pattern Stippling                                                            & 2.7895 & 0.0316 & 0.0085 & 0.0101   & 80.40  \\ 
\midrule
Cube Surface Stippling                                                              & 2.0677 & 0.0387 & 0.0055 & 0.0063   & 82.45  \\ 
\midrule
Cylinder Surface Stippling                                                          & 2.3240 & 0.0282 & 0.0058 & 0.0087   & 80.70  \\
\bottomrule
\end{tabular}
}
\caption{The \textbf{constraint errors (Q3)} and the \textbf{ratio of valid points (VPR (\%))} of different tasks in the real-world.}
\label{tab: appendix-realexp-result}
\vspace{-0.5cm}
\end{table}

\subsection{Generalization and Robustness Capability.} 
To evaluate the generalization capability of our model, we tested the constraint values of the In-Distribution (ID) and Out-of-Distribution (OOD) data under different tasks. In Task \emph{Orientation Alignment}, we set the relative pose to values out of distribution and the different positions of the midpoint together. In Task \emph{Rectangle Constrained Sampling}, we tested the different bounding box sizes. As shown in Table~\ref{tab: compare generalization}, even when we input data from outside the training set, our model can still produce valid sampling results. 

\begin{figure}[]            
  \centering                   
  \includegraphics[width=2.3in]{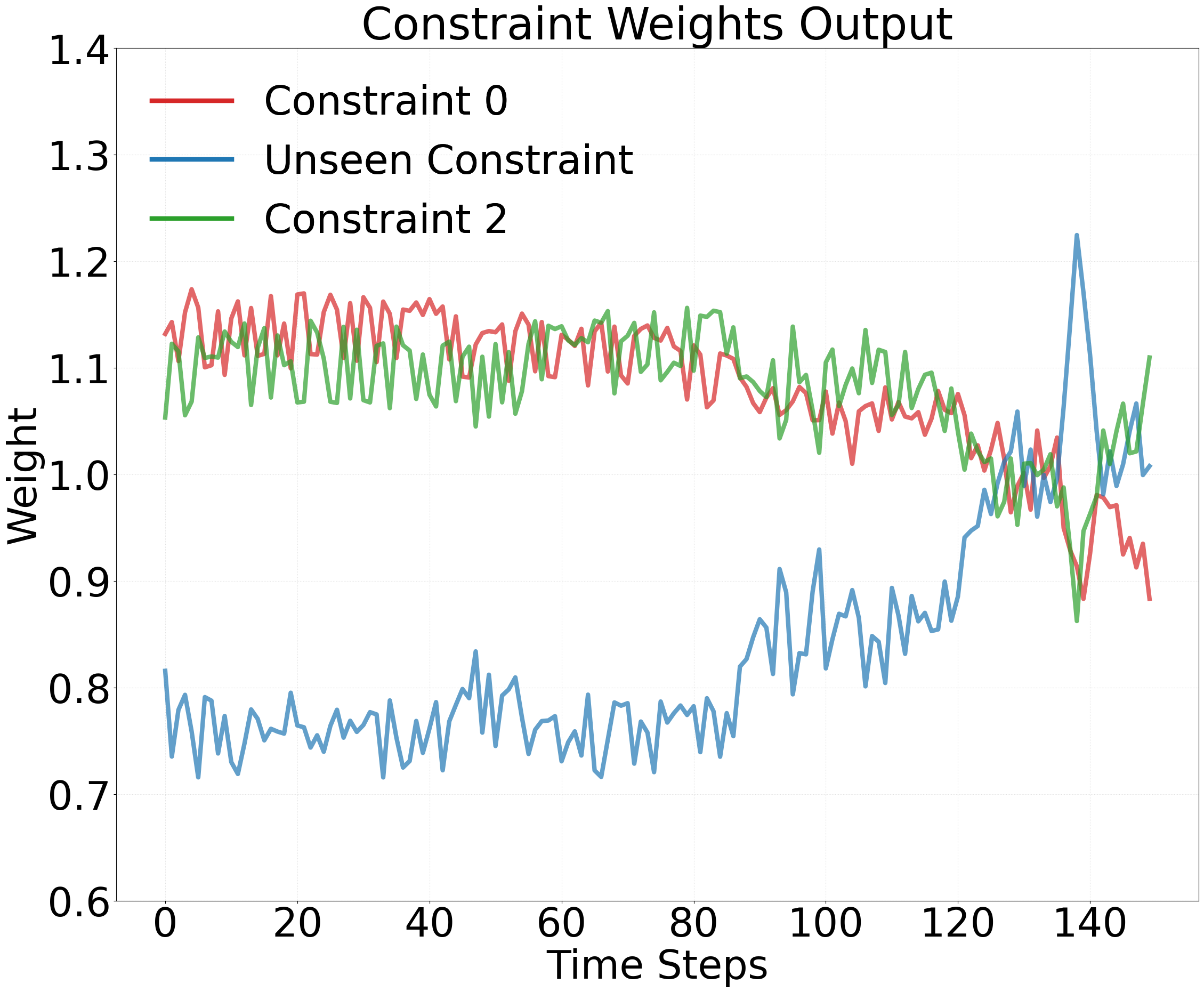}
  \caption{Constraint weight outputs with an \textbf{unseen constraint}.}
  \label{fig:weights_unseen}  
  \vspace{-0.2cm}
\end{figure}

In addition, we evaluated the \textbf{generalization ability of CWT} to unseen constraints. We trained an additional energy network corresponding to an unseen constraint. As illustrated in Figure~\ref{fig:weights_unseen}, CWT is able to assign reasonable weights even to unseen constraints, thereby ensuring that the generated samples remain compliant with the imposed constraints.

\begin{table}
\centering
\begin{tabular}{ccccc} 
\toprule
    & RPE    & MPE    & RRE    & ERE     \\ 
\midrule
\multicolumn{5}{c}{Orientation Alignment}               \\ 
\midrule
ID  & 0.0060 & 0.0029 & 1.6e-5 & -       \\
OOD & 0.0535 & 0.0159 & 9.8e-5 & -       \\ 
\midrule
\multicolumn{5}{c}{Rectangle Constrained Sampling}               \\ 
\midrule
ID  & 0.0002 & 0.2267 & 2.3e-7 & 1.8e-7  \\
OOD & 0.0002 & 0.2775 & 2.3e-7 & 1.9e-7  \\
\bottomrule
\end{tabular}
% \vspace{-0.2cm}
\caption{\textbf{Sampling results of both ID and OOD datasets} to demonstrate the generalization capability. The errors are reported in Q3.}
\label{tab: compare generalization}
\vspace{-0.5cm}
\end{table}

\subsection{Ablation Studies}

We conduct ablations to evaluate the contributions of key components in our method. Specifically, we analyze: (1) The effect of the CWT for dynamic constraint composition. (2) The role of KDE for diversity-aware resampling. (3) The impact of two-stage sampling strategies.

\textbf{Dynamic Weighting via CWT.} \hspace{0.3em} We compare three composition mechanisms for constraint weighting: fixed scalar weights, MLP-based dynamic weights, and our proposed CWT. All models share the same architecture for the energy and feature networks and are trained without task likelihoods to isolate the effect of the weighting mechanism. As shown in Table~\ref{tab:weights_ablation}, CWT consistently outperforms the other approaches, achieving lower errors across all tasks, underscoring its ability to adaptively balance constraint satisfaction in diverse compositions.

\begin{table}[]
    \centering
    \begin{tabular}{ccccc} 
    \toprule
                  & RPE              & MPE              & RRE             & ERE              \\ 
    \midrule
    \multicolumn{5}{c}{ Orientation Alignment}                                                               \\ 
    \midrule
    Fixed Weights & 42.5589          & 43.4433          & 0.0403          & 0.0660           \\
    ~MLP-based~   & 40.9558          & 43.1051          & 0.0403          & 0.0668           \\
    CWT           & \textbf{35.0300} & \textbf{34.7487} & \textbf{0.0337} & \textbf{0.0640}  \\ 
    \midrule
    \multicolumn{5}{c}{Surface Contact Sampling}                                                               \\ 
    \midrule
    Fixed Weights & 63.3622          & 27.1177          & 0.0682          & 0.1195           \\
    ~MLP-based~   & 73.4915          & 27.7264          & 0.0724          & 0.1014           \\
    CWT           & \textbf{62.1286} & \textbf{20.2202} & \textbf{0.0393} & \textbf{0.0857}  \\
    \bottomrule
    \end{tabular}
    \caption{Comparison of constraint composition mechanisms in ADCS: fixed weights, MLP-based dynamic weights, and the proposed CWT. And there is no two-stage sampling involved in this ablation and all variants use the same energy and feature networks and exclude task likelihood information during training. The errors are reported in Q3.}
    \label{tab:weights_ablation}
    \vspace{-0.5cm}
\end{table}

\textbf{KDE-based Resampling.} \hspace{0.3em}
Under Task \emph{Orientation Alignment}, we compared KDE-based resampling against uniform-replication resampling. We estimate the data distribution density with a voxel size of 5. From Table~\ref{tab: kde compare}, we can conclude that the KDE-based resampling method achieves a more uniform data distribution both after the resampling stage and in the final sampling results.

% Under Task 2, we compared KDE-based resampling against uniform-replication resampling, as shown in Figure~\ref{fig:kernel result}, the final samples obtained through Gaussian-kernel replication are also more uniformly distributed. We estimate the data distribution density with voxel size 5. From Table~\ref{tab: kde compare}, we can conclude that the KDE-based resampling method achieves a more uniform data distribution both after the resampling stage and in the final sampling results.

% \begin{figure}
%   \centering
%   \begin{subfigure}[t]{0.48\textwidth}
%     \centering
%     \includegraphics[width=\linewidth,keepaspectratio]{images/uniform_replication_result/Density_obs.png}
%     \caption{Uniform replication sampling result}
%     % \label{fig:task1}
%   \end{subfigure}\hfill
%   \begin{subfigure}[t]{0.48\textwidth}
%     \centering
%     \includegraphics[width=\linewidth,keepaspectratio]{images/gaussian_kernel_replication_result/Density_obs.png}
%     \caption{KDE-based replication sampling result} 
%     % \label{fig:task2}
%   \end{subfigure}\hfill
%   \newline

%   \caption{This figure shows the one end-effector's position distribution densities along the x, y, and z axes of the final sampling results after applying uniform replication and after applying Gaussian-kernel–based replication.}
%   \label{fig:kernel result}
% \end{figure}

\begin{table}[]
\centering
% \resizebox{3in}{!}{
\begin{tabular}{ccc} 
\toprule
\multicolumn{3}{c}{Data Distribution Uniformity (Density Variance)}  \\ 
\midrule
                    & After Replication & Final Sampled              \\ 
Uniform replication & 78.0000           & 79.6028                    \\
KDE replication     & \textbf{50.4320}  & \textbf{57.4560}           \\
\bottomrule
\end{tabular}
% }
% \vspace{0.5em}
\caption{The table presents the variance of the end-effector’s positional \textbf{density distribution for the uniform-replication and KDE-based-replication methods}, both after the resampling stage and after the final sampling. A smaller variance value indicates a more uniform distribution.}
\label{tab: kde compare}
\vspace{-0.5cm}
\end{table}

\textbf{Two-stage Batch-wise Sampling.} \hspace{0.3em} 
To evaluate the performance gains brought by two-stage sampling, we tested ADCS and ADCS without two-stage sampling on three different tasks. And each task uses the same threshold, a sample point is considered valid if its error is smaller than the threshold. As shown in Table~\ref{tab: seq sampling}, for each task, two-stage sampling consistently yields performance improvements, especially on tasks with smaller feasible regions.
% \vspace{-0.5cm}
\begin{table}[H]
\centering
% \resizebox{\textwidth}{!}{
\begin{tabular}{cccc} 
\toprule
\multicolumn{4}{c}{Valid Sample Rate (\%)}                                                                                                                                                                            \\ 
\midrule
              & \begin{tabular}[c]{@{}c@{}}Obstacle \\Avoidance\end{tabular} & \begin{tabular}[c]{@{}c@{}}Orientation \\Alignment\end{tabular} & \begin{tabular}[c]{@{}c@{}}Reactangle \\Constrained Sampling\end{tabular}  \\
w/o Two-stage & 82.3                                                         & 3.74                                                            & 1.54                                                                 \\
Two-stage     & \textbf{99.46}                                               & \textbf{99.24}                                                  & \textbf{99.7}                                                        \\
\bottomrule
\end{tabular}
% }
% \vspace{0.5em}
\caption{Comparison of \textbf{Valid Sample Rates for ADCS with and without two-stage Sampling}.}
\label{tab: seq sampling}
\end{table}

 \textbf{Limitations.}~~~In this work, the SE(3) constraints are defined by manually specifying the corresponding numerical values, rather than through instructions like ``align both hands''. An interesting direction would be to combine this with Vision-Language Models (VLMs) and define the related constraints through Large Language Models (LLMs). Furthermore, LLMs could be employed to define cost functions for the Gauss-Newton refinement, thereby allowing a more general optimization framework. Lastly, the current training process does not incorporate information about the corresponding joint configurations. Future improvements include joint information during the training phase to improve model performance and accelerate convergence during inference.
\section{Conclusion}

In this paper, we introduced ADCS, a generative framework designed to effectively integrate equality and inequality geometric constraints within multi-DoF robot manipulation tasks. Our approach uses a dedicated energy network to calculate the cost of equality and inequality constraints, and introduces an SDF network to incorporate external environment perception, while combined with a Transformer-based architecture for dynamically weighting these constraints during inference. Experimental evaluations across both simulation and real-world tasks demonstrated that ADCS outperforms baseline methods, also applicable to scenarios with small feasible areas and strict spatial constraints. Additionally, through ablation studies, we verified that transformer-based dynamic weights, KDE-based resampling, and batch-wise constrained sampling significantly enhance sample uniformity and performance. Moreover, our sampler can be used to construct a roadmap for enabling motion planning. Finally, our method's robust generalization was validated on out-of-distribution data, confirming its adaptability and efficiency in realistic robotic applications.

\bibliographystyle{IEEEtran}
\bibliography{main}

\end{document}